\pdfoutput=1
\documentclass[conference]{IEEEtran}
\usepackage{graphicx}
\usepackage{CADSLTT}
\usepackage{times}

\newcommand{\overbar}[1]{\mkern 1.7mu\overline{\mkern-1.7mu#1\mkern-1.7mu}\mkern 1.7mu}

\ifCLASSINFOpdf
\else
\fi
\begin{document}
%
% paper title
% can use linebreaks \\ within to get better formatting as desired
\title{Communication-efficient Algorithm for Distributed Sparse Learning via Two-way Truncation}

% author names and affiliations
% use a multiple column layout for up to three different
% affiliations
\author{\IEEEauthorblockN{Jineng Ren and Jarvis Haupt}
\IEEEauthorblockA{Department of Electrical and Computer Engineering\\
University of Minnesota Twin Cities\\
Email: {\tt\{renxx282, jdhaupt\}@umn.edu}}}
%\and
%\IEEEauthorblockN{Xingguo Li}
%\IEEEauthorblockA{Department of Electrical and\\Computer Engineering\\
%	University of Minnesota Twin Cities\\
%	Email: lixx1661@umn.edu}
%\and
%\IEEEauthorblockN{Jarvis Haupt}
%\IEEEauthorblockA{Department of Electrical and\\Computer Engineering\\
%	University of Minnesota Twin Cities\\
%	Email: jdhaupt@umn.edu}}

% conference papers do not typically use \thanks and this command
% is locked out in conference mode. If really needed, such as for
% the acknowledgment of grants, issue a \IEEEoverridecommandlockouts
% after \documentclass

% for over three affiliations, or if they all won't fit within the width
% of the page, use this alternative format:
% 
%\author{\IEEEauthorblockN{Michael Shell\IEEEauthorrefmark{1},
%Homer Simpson\IEEEauthorrefmark{2},
%James Kirk\IEEEauthorrefmark{3}, 
%Montgomery Scott\IEEEauthorrefmark{3} and
%Eldon Tyrell\IEEEauthorrefmark{4}}
%\IEEEauthorblockA{\IEEEauthorrefmark{1}School of Electrical and Computer Engineering\\
%Georgia Institute of Technology,
%Atlanta, Georgia 30332--0250\\ Email: see http://www.michaelshell.org/contact.html}
%\IEEEauthorblockA{\IEEEauthorrefmark{2}Twentieth Century Fox, Springfield, USA\\
%Email: homer@thesimpsons.com}
%\IEEEauthorblockA{\IEEEauthorrefmark{3}Starfleet Academy, San Francisco, California 96678-2391\\
%Telephone: (800) 555--1212, Fax: (888) 555--1212}
%\IEEEauthorblockA{\IEEEauthorrefmark{4}Tyrell Inc., 123 Replicant Street, Los Angeles, California 90210--4321}}

% use for special paper notices
%\IEEEspecialpapernotice{(Invited Paper)}

% make the title area
\maketitle

\begin{abstract}
%\boldmath
We propose a communicationally and computationally efficient algorithm for high-dimensional distributed sparse learning. At each iteration, local machines compute the gradient on local data and the master machine solves one shifted $l_1$ regularized minimization problem. The communication cost is reduced from constant times of the dimension number for the state-of-the-art algorithm to constant times of the sparsity number via Two-way Truncation procedure. Theoretically, we prove that the estimation error of the proposed algorithm decreases exponentially and matches that of the centralized method under mild assumptions. Extensive experiments on both simulated data and real data verify that the proposed algorithm is efficient and has performance comparable with the centralized method on solving high-dimensional sparse learning problems. 
\end{abstract}
% IEEEtran.cls defaults to using nonbold math in the Abstract.
% This preserves the distinction between vectors and scalars. However,
% if the conference you are submitting to favors bold math in the abstract,
% then you can use LaTeX's standard command \boldmath at the very start
% of the abstract to achieve this. Many IEEE journals/conferences frown on
% math in the abstract anyway.

% no keywords

% For peer review papers, you can put extra information on the cover
% page as needed:
% \ifCLASSOPTIONpeerreview
% \begin{center} \bfseries EDICS Category: 3-BBND \end{center}
% \fi
%
% For peerreview papers, this IEEEtran command inserts a page break and
% creates the second title. It will be ignored for other modes.
\IEEEpeerreviewmaketitle

\section{Introduction}
% no \IEEEPARstart
One important problem in machine learning is to find the minimum of the expected loss, 
\begin{align}\label{origin}
\min_{\theta} \mathbb{E}_{\Xb,Y\sim \cD}\left[l(Y,\langle \Xb,\theta \rangle) \right]. 
\end{align}
Here $l(\cdot,\cdot)$ is a loss function and $(\Xb,Y) \in \cX\times \cY \subseteq \mathbb{R}^d \times \cY$ has a distribution $\cD$. In practice, the minimizer $\theta^*$ needs to be estimated by observing $N$ samples $\{\xb_i,y_i\}$ drawn from distribution $\cD$. In many applications $N$ or $d$ are very large, so distributed algorithms are necessary in such case. Without loss of generality, assume that $N = nm$ and that the observations of $j$-th machine are $\{\xb_{ji},y_{ji}\}_{i =1}^n$. We consider the high-dimensional learning problem where the dimension $d$ can be very large, and the effective variables are supported on $S:= \text{support}\{\theta^*\} = \{i\in [d]:\theta^*_i \ne 0\}$ and $s:=|S|\ll d$. Extensive efforts have been made to develop batch algorithms \cite{friedman2007pathwise,Beck:09,xiao2013proximal}, which provide good convergence guarantees in optimization. However, when $N$ is large, batch algorithms are inefficiency, which takes at least $\cO(N)$ time per iteration. Therefore, an emerging recent interest is observed to address this problem using the distributed optimization frameworks \cite{jordan2016communication,lee2015distributed,wang2016efficient}, which is more efficient than the stochastic algorithms. One important issue of existing distributed optimization for sparse learning is that they did not take advantage of the sparse structure, thus they have the same communication costs with general dense problems. In this paper, we propose a novel communication-efficient distributed algorithm to explicitly leverage the sparse structure for solving large scale sparse learning problems. This allows us to reduce the communication cost from $\cO(d)$ in existing works to $\cO(s)$, while we still maintaining nearly the same performance under mild assumptions.

\noindent \textbf{Notations}
For a sequence of numbers $a_n$, we use $\cO(a_n)$ to denote a sequence of numbers $b_n$ such that $b_n\leq C \cdot a_n$ for some positive constant $C$. Given two sequences of numbers $a_n$ and $b_n$, we say $a_n\lesssim b_n$ if $a_n =\cO(b_n) $ and $a_n\gtrsim b_n$ if $b_n =\cO(a_n) $. The notation $a_n \asymp b_n$ denotes that $a_n =\cO(b_n) $ and $b_n =\cO(a_n) $. For a vector $\vb\in \mathbb{R}^d$, the $l_p$-norm of $\vb$ is defined as $\|\vb\|_p = ( \sum_{i =1}^d|\vb_i|^p)^{1/p}$, where $p>0$; the $l_0$-norm of $\vb$ is defined as the number of its nonzero entries; the support of $\vb$ is defined as $\text{supp}(\vb) = \{i:\vb_i\ne 0 \}$. For simplicity, we use $[d]$ to denote the set $\{1,\cdots,d\}$. For a matrix $A = (a_{ij})\in \mathbb{R}^{n_1\times n_2}$, we define the $l_\infty$-norm of $A$ as $\|A\|_\infty = \max_{i\in[n_1],j\in[n_2]}|a_{ij}|$. Given a number $k\leq d$, the hard thresholding $\cH_k(\vb)$ of a vector $\vb\in\RR^d $ is defined by keeping the largest $k$ entries of $\vb$ (in magnitude) and setting the rest to be zero. Given a subset $S$ of index set $\{1,\cdots,d\}$, the projection $\cP_S(\vb)$ of a vector $\vb$ on $S$ is defined by 
\begin{align*} 
\cP_S(\vb)_j =0,\hspace{0.1in} \text{if} \hspace{0.05in} j\notin S \hspace{0.1in} \text{and}\hspace{0.1in} \cP_S(\vb)_j =\vb_j, \hspace{0.1in}\text{if}\hspace{0.05in} j\in S.
\end{align*} 
$\cP_S(\vb)$ is also denoted as $(\vb)_S$ for short.

\vspace{-0.1in}
\subsection{{Related work}}
There is much previous work on distributed optimizations such as (Zinkevich et al. \cite{zinkevich2010parallelized}; Dekel et al. \cite{dekel2012optimal}; Zhang et al. \cite{zhang2012communication}; Shamir and Srebro \cite{shamir2014distributed}; Arjevani and Shamir \cite{arjevani2015communication}; Lee et al. \cite{lee2015distributed}; Zhang and Xiao \cite{zhang2015disco}). Initially, most distributed algorithms used averaging estimators formed by local machines (Zinkevich et al. \cite{zinkevich2010parallelized}; Zhang et al. \cite{zhang2012communication}). Then Zhang and Xiao \cite{zhang2015disco}, Shamir et al. \cite{shamir2014communication} and Lee et al. \cite{lee2015communication} proposed more communication-efficient distributed optimization algorithms. More recently, using ideas of the approximate Newton-type method, Jordan et al. \cite{jordan2016communication} and Wang et al. \cite{wang2016efficient} further improved the computational efficiency of this type of method.

Many gradient hard thresholding approaches are proposed in recent years such as (Yuan et al. \cite{yuan2014gradient}; Li et al. \cite{li2016stochastic}; Jain et al. \cite{jain2014iterative}). They showed that under suitable conditions, the hard thresholding type first-order algorithms attain linear convergence to a solution which has optimal estimation accuracy with high probability. However, to the best of our knowledge, hard thresholding techniques applied to approximate Newton-type distributed algorithms has not been considered yet.  So in this paper, we present some initial theoretical and experimental results on this topic. 
\iffalse
Initially, averaging estimators formed by locally machines is a intuitive approach to distributed estimation (Zinkevich et al. \cite{zinkevich2010parallelized}; Zhang et al. \cite{zhang2012communication}).
\fi
\vspace{-0.15in}

\section{Algorithm} 
\vspace{-0.05in}
In this section, we explain our approach to estimating the $\theta^*$ that minimizes the expected loss. The detailed steps are summarized in Algorithm \ref{algor}. 

First the empirical loss at each machine is defined as
\begin{align*}\textstyle
\cL_j(\theta) = \frac{1}{n} \sum_{i =1}^n l(y_{ji},\langle \xb_{ji},\theta \rangle), ~~\text{where}~~ j\in [m]. 
\end{align*}
At the beginning of algorithm, we solve a local Lasso subproblem to get an initial point. Specifically, at iteration $h =0$, the master machine solves the minimization problem
\begin{align} \label{initial}
\textstyle \gamma^0 = \argmin \cL_1(\theta) + \mu_0\|\theta\|_1. 
\end{align}
The initial point $\theta^0$ is formed by keeping the largest $k$ elements of the resulting minimizer $\gamma^0$ and setting the other elements to be zero, i.e., $\theta^0 = \cH_k(\gamma^0)$. Then, $\theta^0$ is broadcasted to the local machines, where it is used to compute a gradient of local empirical loss at $\theta^0$, that is, $\nabla\cL_j(\theta^0)$. The local machines project $\nabla\cL_j(\theta^0)$ on the support $S^0$ of $\theta^0$ and transmit the projection $\cP_{S^0}\left[\nabla\cL_j(\theta^0)\right]$ back to the master machine. Later at $(h+1)$-th iteration ($h\geq 0$), the master solves a shifted $l_1$ regularized minimization subproblem:
\begin{align}\label{subproblem}
\textstyle \nonumber \gamma^{h+1}& = \argmin_{ \theta} \hspace{0.1in}\cL_1(\theta) + \mu_{h+1} \|\theta\|_1\\
&+\Big\langle \cP_{S^h}\left[{\textstyle\frac{1}{m}\sum_{j=1}^m \nabla \cL_j(\theta^h)}\right] -\nabla \cL_1(\theta^h),\theta \Big \rangle.  
\end{align} 
Again the minimizer $\gamma^{h+1}$ is truncated to form $\theta^{h+1}$, and this quantity is communicated to the local machines, where it is used to compute the local gradient as before. 

Solving subproblem $\eqref{subproblem}$ is inspired by the approach of Wang et al. \cite{wang2016efficient} and Jordan et al.\cite{jordan2016communication}. Note that the formulation takes advantage of both global first-order information and local higher-order information. Specially, assuming the $\mu_{h+1} = 0$ and $\cL_j$ has an invertible Hessian, the solution of $\eqref{subproblem}$ has the following closed form
\begin{align*}
\gamma^{h+1} = \theta^{h} - \nabla^2\cL_1(\theta^h) ^{-1} \left(\cP_{S^h}\left[{\textstyle\frac{1}{m}\sum_{j=1}^m \nabla \cL_j(\theta^h)}\right]\right),
\end{align*}
which is similar to a Newton updating step. Note that here we add a projection procedure $\cP_{S^h}\left[\frac{1}{m}\sum_{j=1}^m\nabla\cL_j(\theta^h)\right]$ to reduce the number of nonzeros that need to be communicated to the master machine. This procedure is reasonable intuitively. First, when $\theta^h$ is close to $\theta^*$, the elements of $\frac{1}{m}\sum_{j=1}^m\nabla \cL_j(\theta^h)$ outside the support $S^{h}$ should be very small, so nominally little error is incurred in the truncation step. Second, when $\theta^{h+1}$ is also close to $\theta^*$, the lost part has even more minimal effects on the inner product in subproblem $\eqref{subproblem}$. Third, we leave $-\nabla \cL_1(\theta^h)$ in $\eqref{subproblem}$ out of the truncation to maintain the formulation as unbiased.

\begin{algorithm}[!htb]
	\caption{Two-way Truncation Distributed Sparse Learning}
	\label{algor}
	%	\label{alg:main1}
	\begin{algorithmic}
		\REQUIRE Loss function $l(\cdot,\cdot)$, data $\{\xb_{ji}, y_{ji}\}_{i\in[n],j\in[m]}$. 
		\STATE \hspace{-1.25em} \textbf{\underline{Local machines:}}
		\STATE \hspace{-1.25em} \textbf{Initializaiton:} The master solves the local $l_1$ regularized loss minimization problem \eqref{initial} to get a solution $\gamma^0$. Set $\theta^{0} = \cH_k(\gamma^{0})$. 
		\FOR{$h=0,1, \dots$}
		\FOR{$j=2,3, \dots, m$}
		\STATE   \textbf{if} Receive $\theta^h$ from the master \textbf{then}
		\STATE Calculate gradient $\nabla \cL_j(\theta^h)$ and get the projection $\cP_{S^{h}}\left[ \nabla\cL_j(\theta^h)\right]$ of the gradient on support $S^h$ and transmit it to the master.
		\STATE \textbf{end}
		
		\ENDFOR
		\STATE  \textbf{\underline{Master:}}
		\STATE   \textbf{if} Receive $\{\nabla\cL_j(\theta^h)\}_{j=2}^m$ from local machines \textbf{then}
		\STATE \hspace{1.25em}Solve the shifted $l_1$ regularized problem
		\STATE \hspace{1.25em}\eqref{subproblem} to obtain $\gamma^{h+1}$.
		\STATE \hspace{1.25em}Do hard thresholding $\theta^{h+1} =         \cH_k(\gamma^{h+1})$. 
		\STATE \hspace{1.25em}Let $S^{h+1} = \text{supp}(\theta^{h+1})$. 
		\STATE \hspace{1.25em}Broadcast $\theta^{h+1}$ to every local machine. 
		\STATE \textbf{end}	
		\ENDFOR	
	\end{algorithmic}
\end{algorithm}
\vspace{-0.1in}
\section{Theoretical Analysis}
\vspace{-0.05in}
\subsection{Main Theorem}

We present some theoretical analysis of the proposed algorithm in this section.
\vspace{-0.05in}
\begin{assumption}\label{assum_smooth}
	The loss $l(\cdot,\cdot)$ is a $L$-smooth function of the second argument, i.e.,
	\begin{align*} {\textstyle
	l^\prime(x,y) - l^\prime(x,z) \leq L|y - z|, \hspace{0.2in} \forall x,y,z\in \RR}
	\end{align*}
	Moreover, the third derivative with respect to its second argument, $\partial^3 l(x,y)/\partial y^3$, is bounded by a constant $M$, i.e.,
	{\begin{align*} \textstyle
	|\partial^3 l(x,y)/\partial y^3| \leq M, \hspace{0.2in} \forall x,y\in \RR
	\end{align*}}
\end{assumption}

\begin{assumption}\label{assum_restrict}
	The empirical loss function computed on the first machine satisfies that: $\forall \Delta \in \cC(S,3)$, we have
	\begin{align*}
	\cL_1(\theta^* + \Delta) - \cL_1(\theta^*) -\langle \nabla \cL_1(\theta^*),\Delta \rangle \geq \kappa \|\Delta\|_2^2,
	\end{align*}
	where $\cC(S,3)$ is defined as
	\begin{align*}
	\cC(S,3) = \{\Delta \in \RR^d |~ \|\Delta_{S^c}\|_1 \leq 3\|\Delta_S\|_1\}. 
	\end{align*}	
\end{assumption}

\begin{assumption}\label{assum_supp}
	The $\gamma^{h+1}$, $S^{h+1}$ and $S^h$ defined in Algorithm \ref{algor} satisfy the following condition:
	there exists some positive constants $H$ and $\tau_1$ and $\tau_2$ such that for $h\geq H$, 
	\begin{align*}
	\left\|\left(\gamma^{h} - \theta^*\right)_{(S^h)^c}\right\|_1 &\leq \tau_1 \left\|\gamma^{h} - \theta^*\right\|_1\\
	\left\|\left(\gamma^{h+1} - \theta^*\right)_{S^{h+1}\backslash S^{h}}\right\|_1 &\leq \tau_2\left\|\gamma^{h+1}-\theta^*\right\|_1.   
	\end{align*}
\end{assumption}
\begin{remark}
	In practice, both $\tau_1$ and $\tau_2$ are very small even after only one round of communication and will decrease to $0$ fast in the later steps. 
\end{remark}
\vspace{-0.1in}
For simplicity, we define the following notation:
\begin{align*}
\overbar{\cL_1}(\theta^*,\theta^h) &:= \cL_1(\theta^*) +\left\langle {\textstyle \frac{1}{m}\sum_{j = 1}^m  \nabla\cL_j(\theta^h) - \nabla\cL_1(\theta^h)},\theta \right \rangle,\\
\tilde{\cL_1}(\theta^*,\theta^h) &:= \cL_1(\theta^*)\\
&\hspace{0.2in}+\left\langle \cP_{S^h}\left[{\textstyle \frac{1}{m}\sum_{j = 1}^m  \nabla\cL_j(\theta^h) - \nabla\cL_1(\theta^h)}\right],\theta \right \rangle.
\end{align*}

Now we state our main theorem. 
\begin{thmi} \label{main_theorem}
	Suppose that Assumption~\ref{assum_smooth}, \ref{assum_restrict}, and \ref{assum_supp} hold. Let $k = C_1\cdot s$ with $C_1>1$ and 
	\begin{align}
	\nonumber& \textstyle\mu_{h+1}  = \hspace{0.1in} 4\left\|{ \frac{1}{m}\sum_{j = 1}^m  \nabla \cL_j(\theta^*)}\right\|_\infty \\
	\nonumber&\textstyle +  2L\left(\max_{j,i}\|x_{j,i}\|_\infty^2\right)\cdot \Big[{ 2\sqrt{\frac{\log(2d/\delta)}{n}} + \rho }\Big]\|\theta^h - \theta^*\|_1\\
	&\textstyle + 2M\left(\max_{j,i}\|x_{j,i}\|_\infty^3 \right)\|\theta^h - \theta^*\|_1^2, \label{mu_def}
	\end{align} 
	where $\rho := \tau_1+\tau_2$.  
	
	Then with probability at least $1-\delta$, we have that
	
	\begin{align*} 
    &\|\theta^{h+1} - \theta^*\|_1  {\textstyle \leq \frac{C_2 s}{\kappa}\left \|\frac{1}{m}\sum_{j=1}^m \nabla \cL_j(\theta^*)\right \|_\infty}\\
    &\hspace{0.2in}{\textstyle +\frac{C_2 s}{2\kappa} L\cdot \max_{j,i}\|\xb_{ji}\|_\infty^2\cdot \Big[2\sqrt{\frac{\log(2d/\delta)}{n}} + \rho \Big]\|\theta^h - \theta^*\|_1}\\
    &\hspace{0.2in}{\textstyle +\frac{C_2 s}{2\kappa}  M\cdot  \max_{j,i}\|\xb_{ji}\|_\infty^3\cdot \|\theta^h - \theta^*\|_1^2}, ~\text{and}\\
    &\|\theta^{h+1} - \theta^*\|_2 {\textstyle \leq \frac{C_3\sqrt{s}}{\kappa}\left \|\frac{1}{m}\sum_{j=1}^m \nabla \cL_j(\theta^*)\right \|_\infty }\\
    &\hspace{0.2in} {\textstyle + \frac{C_3\sqrt{s}}{2\kappa} L \cdot \max_{j,i}\|\xb_{ji}\|_\infty^2\cdot\Big[2\sqrt{\frac{\log(2d/\delta)}{n}} + \rho \Big]\cdot \|\theta^h - \theta^*\|_1} \\
    &\hspace{0.2in} \textstyle+\frac{C_3\sqrt{s}}{2\kappa}M\cdot \max_{j,i}\|\xb_{ji}\|_\infty^3\cdot \|\theta^h - \theta^*\|_1^2, 
    \end{align*}
	where $\textstyle C_2 = 24\sqrt{1+2(C_1-1)^{-\frac{1}{2}}}\cdot \sqrt{C_1+1}$ and $\textstyle C_3= 24\sqrt{1+2(C_1-1)^{-\frac{1}{2}}}$ are positive constants independent of $m,n,s,d$.
\end{thmi}
The theorem immediately implies the following convergence result.
\begin{cori} \label{first_cor}
	Suppose that for all $h$
	\begin{align}\label{cor_ass}
	\textstyle \nonumber&M\cdot  \Big(\max_{j,i}\|x_{ji}\|_\infty^3\Big) \|\theta^h - \theta^*\|_1\leq\\ &  L\cdot \max_{j,i}\|x_{ji}\|_\infty^2\left[2\sqrt{\frac{\log(2d/\delta)}{n}} 
	+ \rho\right],
	\end{align}
	where $\rho := \tau_1+\tau_2$.
	
	Then under the assumption of Theorem~\ref{main_theorem} we have
	\begin{align*} 
	\|\theta^{h+1} - \theta^*\|_1 &\leq \textstyle \frac{1-a_n^{h+1}}{1 - a_n}\cdot \frac{C_2 s}{\kappa}\cdot \textstyle \left\|\frac{1}{m}\sum_{j=1}^m \nabla \cL_j(\theta^*)\right\|_\infty \\
	& \hspace{0.1in}+ a_n^{h +1} \|\theta^0 - \theta^*\|_1,\\
	|\theta^{h+1} - \theta^*\|_2 &\leq \textstyle \frac{1-a_n^{h+1}}{1 - a_n}\cdot \frac{C_3\sqrt{s}}{\kappa}\cdot \textstyle \left\|\frac{1}{m}\sum_{j=1}^m \nabla \cL_j(\theta^*)\right\|_\infty \\
	&\hspace{0.1in}+ a_n^h b_n  \|\theta^0 - \theta^*\|_1,
	\end{align*}
	where 
   \begin{align*}
	 a_n =&\textstyle \frac{C_2 s}{\kappa} L\cdot \max_{j,i}\|x_{ji}\|_\infty^2 \cdot \Bigg[2\sqrt{\frac{\log(2d/\delta)}{n}}+ \rho\Bigg]
	\end{align*}
	and 
	\begin{align*}
	 b_n =&\textstyle \frac{C_3\sqrt{s}}{\kappa} L\cdot \max_{j,i}\|x_{ji}\|_\infty^2\cdot\Bigg[ 2\sqrt{\frac{\log(2d/\delta)}{n}}+ \rho \Bigg],
	\end{align*}
	where $C_2 $ and $C_3$ are defined in Theorem~\ref{main_theorem} and independent of $m,n,s,d$. 
\end{cori}
\begin{remark}
	From the conclusion, we know that the hard thresholding parameter $k$ can be chosen as $C_1\cdot s$, where $C_1$ can be a moderate constant larger than $1$. By contrast, previous work such as \cite{li2016stochastic} solving a nonconvex minimization problem subject to $l_0$ constraint $\|\theta\|_0 \leq k$ requires that $k\geq \cO(\kappa_s^2 s)$, where $\kappa_s$ is the condition number of the object function.
	Moreover, instead of only hard thresholding on the solution of Lasso subproblems, we also do projection on the gradients in \eqref{subproblem}. These help us reduce the communication cost from $\cO(d)$ to $\cO(s)$. 
\end{remark}
%\begin{remark}
%	From the conclusion, we know that the hard thresholding parameter $k$ can be chosen as $C_1\cdot s$, where $C_1$ can be a moderate constant larger than $1$. 
%\end{remark}

\subsection{Sparse Linear Regression}
In the sparse linear regression, data $\{\xb_{ji}, y_{ji}\}_{i\in [n], j\in[m]}$ are generated
according to the model
\begin{align} \label{lin_model} 
\textstyle y_{ji} = \langle\xb_{ji},\theta^*\rangle + \epsilon_{ji},
\end{align}
where the noise $\epsilon_{ji}$ are i.i.d subgaussian random variables with zero mean. Usually the the loss function for this problem is the squared loss function $l(y_{ji},\langle \theta,\xb_{ji}\rangle) = \frac{1}{2}(y_{ji}-\langle \theta,\xb_{ji}\rangle)^2$, which is $1$-smooth.

Combining Corollary \ref{first_cor} with some intermediate results obtained from \cite{rudelson2011reconstruction, vershynin2010introduction} and \cite{wainwright2009sharp}, we have the following bound for the estimation error. 
\begin{cori}
	Suppose the design matrix and noise are subgaussian, Assumption \ref{assum_supp} holds and $\mu_{h+1}$ is defined as \eqref{mu_def}. Then under the sparse linear model, we have the following estimation error bounds with probability at least $1-2\delta$:
	\begin{align*} \textstyle
	\|\theta^{h+1} - \theta^*\|_1 \lesssim \frac{1-a_n^{h+1}}{1-a_n} \cdot\frac{C_2s\sigma\sigma_X}{\kappa}\sqrt{\frac{\log(d/\delta)}{mn}}\\
	\textstyle + a_n^{h+1} \frac{s \sigma \sigma_X}{\kappa}\sqrt{\frac{\log(nd/\delta)}{n}}
	\end{align*}
	and
	\begin{align*} \textstyle
	\|\theta^{h+1} - \theta^*\|_2 \lesssim\frac{1-a_n^{h+1}}{1-a_n} \cdot\frac{C_3\sqrt{s}\sigma \sigma_X}{\kappa}\sqrt{\frac{\log(d/\delta)}{mn}}\\
	\textstyle + a_n^{h}b_n \frac{s \sigma \sigma_X}{\kappa}\sqrt{\frac{\log(nd/\delta)}{n}},
	\end{align*}
	where $C_2 $ and $ C_3$ are defined in Theorem~\ref{main_theorem}, and where
	\begin{align*} 
	\textstyle a_n =\frac{C_2s}{\kappa}\sigma_X^2\log\left(\frac{mnd}{\delta}\right)\left[2\sqrt{\frac{\log(2d/\delta)}{n}} + \rho \right] 
	\end{align*}
	and
	\begin{align*} 
	\textstyle b_n = \frac{C_3\sqrt{s}}{\kappa}\sigma_X^2\log\left(\frac{mnd}{\delta}\right)\left[2\sqrt{\frac{\log(2d/\delta)}{n}} + \rho \right]. 
	\end{align*}	
\end{cori}

\begin{remark}
Under certain conditions we can further simplify the bound and have an insight of the relation between $n,m,s,d$. When $n\geq s^2\log d$, it is easy to see by choosing
\begin{align*}
\mu_{h+1} \asymp \sqrt{\frac{\log d}{mn}} + \sqrt{\frac{\log d}{n}} \left[s\left( \sqrt{\frac{\log d}{n}} + \rho \right)\right]^{h+1} 
\end{align*}
and $k = \cO(s)$ there holds the following error bounds with high probabiltiy:
\begin{align*}
&\|\theta^{h+1} - \theta^*\|_1 \\
&\hspace{0.2in} \lesssim s\sqrt{\frac{\log d}{mn}} + s\sqrt{\frac{\log d}{n}} \left[s\left( \sqrt{\frac{\log d}{n}} + \rho \right)\right]^{h+1},\\
&\|\theta^{h+1} - \theta^*\|_2 \\
&\hspace{0.2in}\lesssim \sqrt{\frac{s\log d}{mn}} +\sqrt{\frac{s\log d}{n}} \left[s\left( \sqrt{\frac{\log d}{n}} + \rho \right)\right]^{h+1}.
\end{align*}
\end{remark}

\subsection{Sparse Logistic Regression}
Combining Corollary \ref{first_cor} with some intermediate results obtained from \cite{wang2016efficient} and \cite{regularizers2012m}, we now can give a similar result about the estimation error bound for sparse logistic regression. The explicit form is omitted due to the limitation of spaces.
\iffalse
\begin{cori}
	Suppose Assumption \ref{assum_supp} holds and $\mu_{h+1}$ is defined as \eqref{mu_def}.
	If the following condition holds for some $T\geq 0$:
	\begin{align*}
	\|\theta^T - \theta^*\|_1 \leq 4\sqrt{\frac{\log (2d/\delta)}{n}}. 
	\end{align*}
	Then under the sparse logistic model with random design, we have the following estimation error bound for all $h\geq T$ with probability at least $1-2\delta$:	
	\begin{align*} \textstyle
	\|\theta^{h+1}- \theta^*\|_1 \lesssim\sqrt{1+2(C_1-1)^{-\frac{1}{2}}}\frac{1-a_n^{h-T+1}}{1-a_n}\hspace{0.27in}\\
	\cdot\frac{24C_2s \sigma_X}{\kappa} \sqrt{\frac{\log(d/\delta)}{mn}}+4a_n^{h-T+1}\sqrt{\frac{\log(2d/\delta)}{n}}
	\end{align*}
	and
	\begin{align*} \textstyle
	\|\theta^{h+1}- \theta^*\|_2 \lesssim \sqrt{1+2(C_1-1)^{-\frac{1}{2}}}\frac{1-a_n^{h-T+1}}{1-a_n}\hspace{0.27in}\\
	\cdot \frac{24\sqrt{s} \sigma_X}{\kappa}\sqrt{\frac{\log(d/\delta)}{mn}} +4a_n^{h-T}\sqrt{\frac{\log(2d/\delta)}{n}},
	\end{align*}
	where
	\begin{align*} \textstyle
	a_n = 6\sqrt{1+2(C_1-1)^{-\frac{1}{2}}}\hspace{1.5in}\\
	\cdot\frac{ C_2 s}{\kappa}\sigma_X^2\log(\frac{mnd}{\delta})\left[\sqrt{\frac{\log(2d/\delta)}{n}} + \rho\right]
	\end{align*}
	and
	\begin{align*} \textstyle
	b_n = 6\sqrt{1+2(C_1-1)^{-\frac{1}{2}}}\hspace{1.5in}\\
	\cdot\frac{\sqrt{s}}{\kappa}\sigma_X^2\log(\frac{mnd}{\delta})\left[\sqrt{\frac{\log(2d/\delta)}{n}} + \rho\right]. 
	\end{align*}
\end{cori}
\fi

\section{Experiments}
\vspace{-0.02in}
Now we test our algorithm on both simulated data and real data. In both settings, we compare our algorithm with various advanced algorithms. These algorithms are:

\begin{itemize}
	\item[1.] EDSL: the state-of-the-art approach proposed by Jialei Wang et al. \cite{wang2016efficient}.
	\vspace{-0.05in}
	\item[2.] Centralize: using all data, one machine solves the centralized loss minimization problem with $l_1$ regularization. This procedure is communication expensive or requires much larger storage.
	\vspace{-0.05in} 
	\item[3.] Local: the first machine solves the local $l_1$ regularized loss minimization problem with only the data stored on this machine, ignoring all the other data.
	\vspace{-0.05in}
	\item[4.] Two-way Truncation: the proposed sparse learning approach which further improves the communication efficiency. 	
	\vspace{-0.1in}
\end{itemize}
\subsection{Simulated data}
%\vspace{-0.1in}
\begin{figure}[!htb]
	\centering 
	\vspace{-0.1in}
	\begin{center}
		%		\hspace{-0.08in}\subfloat[m =5]{
		%			\includegraphics[width=0.32\linewidth]{{simulated_data_good_result/200_1000_10_1_5}.pdf}
		%		}
		%		\hspace{-0.08in}\subfloat[m =5]{
		%					\includegraphics[width=0.32\linewidth]{{simulated_data_good_result/600_2000_10_1_5}.pdf}
		%		}
		\hspace{-0.096in}\subfloat[$\Sigma_{ij} = 0.5^{|i -j|}$]{
			\includegraphics[width=0.46\linewidth]{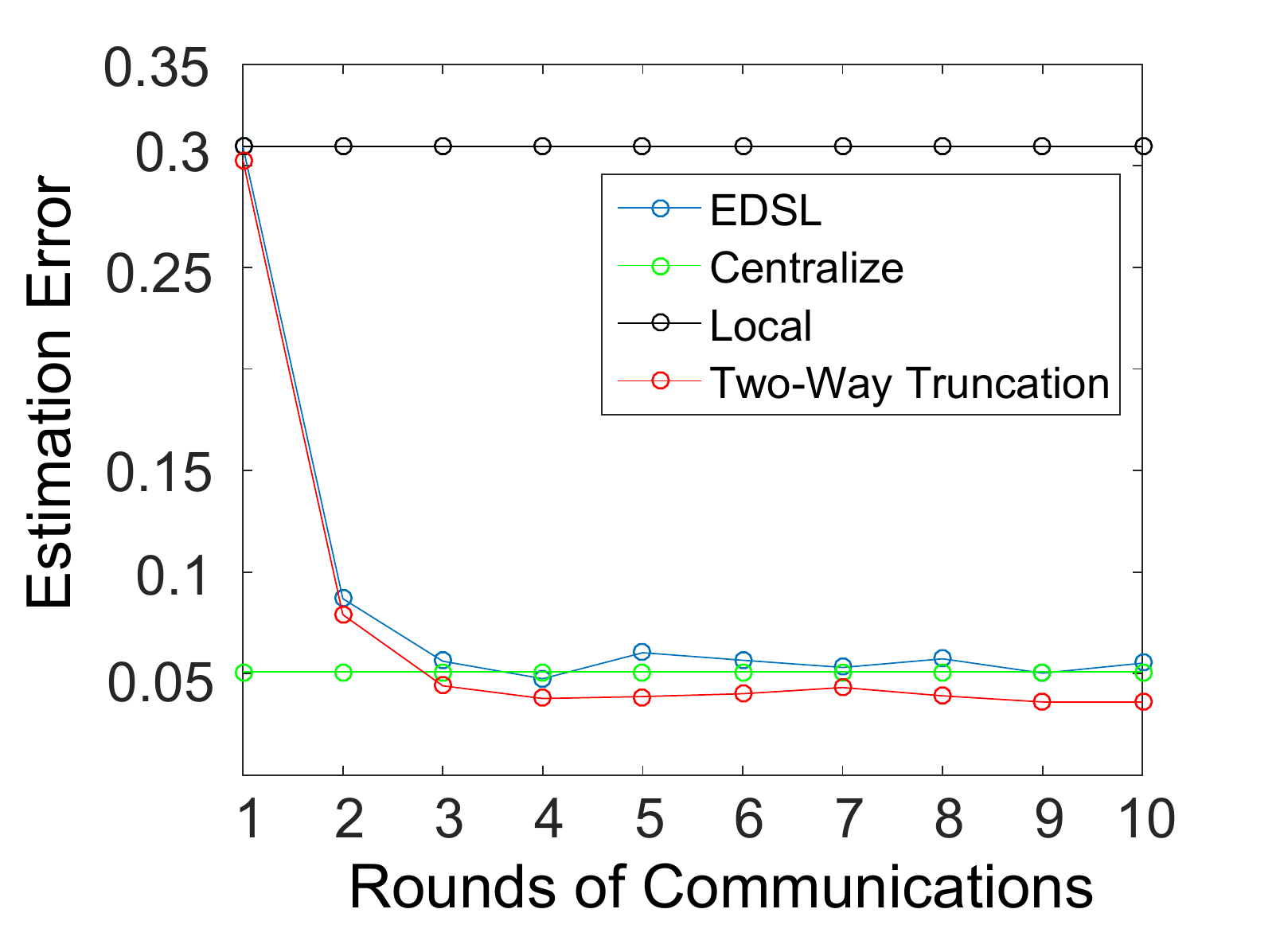}
		}
		\hspace{-0.08in}\subfloat[$\Sigma_{ij} = 0.5^{|i -j|/5}$]{
			\includegraphics[width=0.46\linewidth]{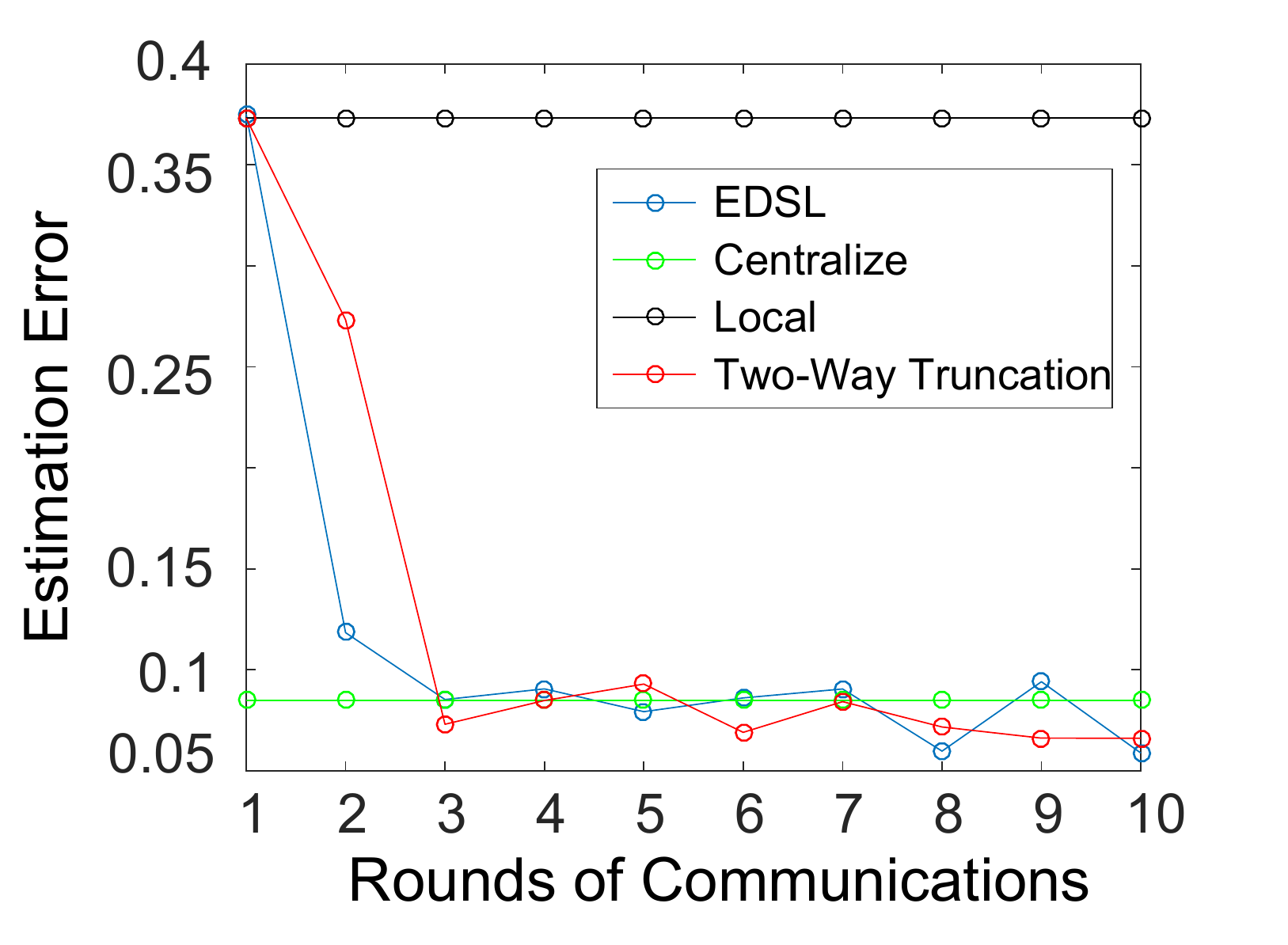}
		}\\
		\vspace{0.1in} $m= 20, n =600, d = 20000, s=10, \Xb \sim \cN(0,\Sigma)$ \vspace{0.1in}\\
		%		\hspace{-0.08in}\subfloat[m =5]{
		%			\includegraphics[width=0.32\linewidth]{{simulated_data_good_result/600_2000_10_0_5}.pdf}
		%		}
		%		\hspace{-0.096in}\subfloat[OP]{
		%			\includegraphics[width=0.45\linewidth]{{plot/OP_outlier_PaviaU}.pdf}
		%		}\hspace{-0.11in}
	\end{center}
	\vspace{-0.1in}	
	 \caption{Comparison among four algorithms in sparse linear regression setting}\label{figure1}
	%\vspace{-0.18in}
\end{figure}

\begin{figure}[!htb] 
	%	\vspace{0.3in}
	\centering
	\begin{center}
		%		\hspace{-0.08in}\subfloat[m =5]{
		%			\includegraphics[width=0.32\linewidth]{{simulated_data_good_result/500_1000_10_1_5_log}.pdf}
		%		}
		%		\hspace{-0.08in}\subfloat[m =5]{
		%			\includegraphics[width=0.32\linewidth]{{simulated_data_good_result/500_1000_10_0_5_log}.pdf}
		%		}
		%		\hspace{-0.08in}\subfloat[m =5]{
		%			\includegraphics[width=0.32\linewidth]{{simulated_data_good_result/1000_2000_10_1_5_log}.pdf}
		%		}
		\hspace{-0.096in}\subfloat[$\Sigma_{ij} = 0.5^{|i -j|}$]{
			\includegraphics[width=0.46\linewidth]{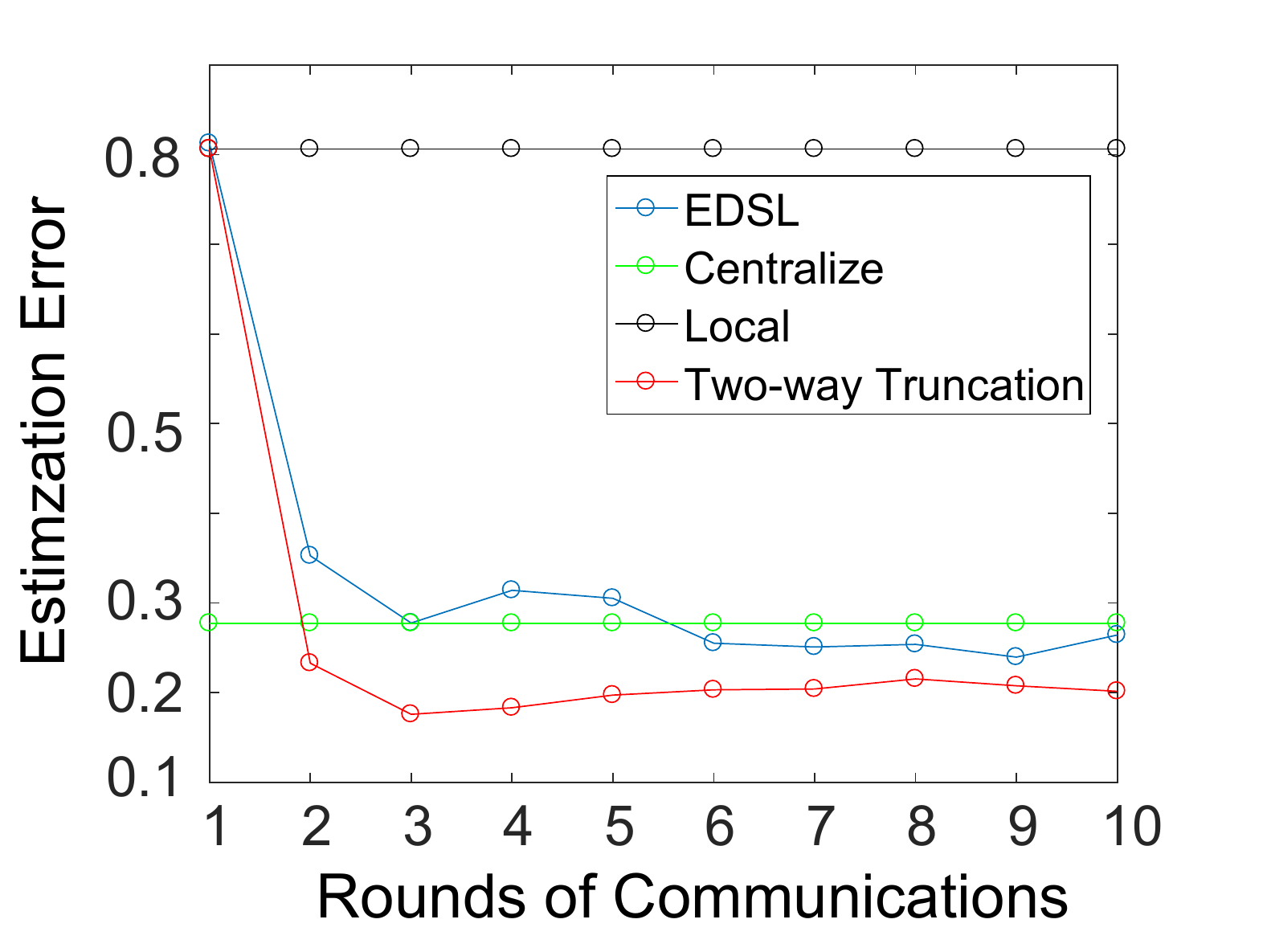}
		}
		\hspace{-0.08in}\subfloat[$\Sigma_{ij} = 0.5^{|i -j|/5}$]{
			\includegraphics[width=0.46\linewidth]{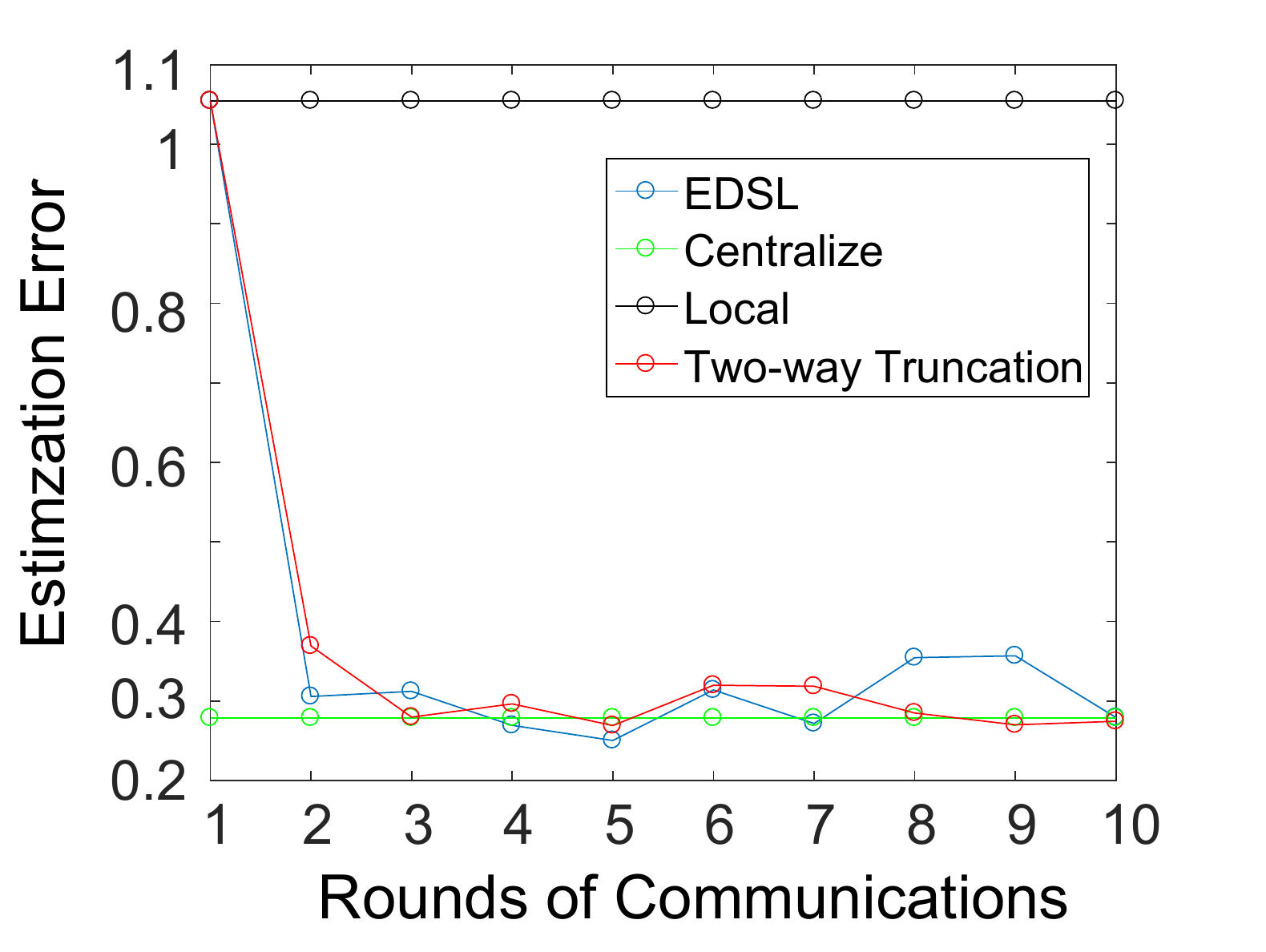}
		}\\
		\vspace{0.1in} $m= 10, n =1000, d = 2000, s=20, \Xb \sim \cN(0,\Sigma)$ \vspace{0.1in}\\
		%		\hspace{-0.08in}\subfloat[m =5]{
		%			\includegraphics[width=0.32\linewidth]{{simulated_data_good_result/1000_2000_10_0_5_log}.pdf}
		%		}
		%		\hspace{-0.096in}\subfloat[OP]{
		%			\includegraphics[width=0.45\linewidth]{{plot/OP_outlier_PaviaU}.pdf}
		%		}\hspace{-0.11in}
	\end{center}
	\vspace{-0.1in}
	\caption{\hspace{-0.1in} Comparison among four algorithms in sparse logistic regression setting}\label{figure2}
	%\vspace{-0.18in}
\end{figure}
The simulated data $\{\xb_{ji}\}_{i\in [n], j\in[m]}$ is sampled from multivariate Gaussian distribution with zero mean and covariance matrix $\Sigma$. We choose two different covariance matrices: $|\Sigma_{ij}| = 0.5^{|i-j|}$ for a well-conditioned situation and $|\Sigma_{ij}| = 0.5^{|i-j|/5}$ for an ill-conditioned situation. The noise $\epsilon_{ji}$ in sparse linear model ($y_{ji} = \langle\xb_{ji},\theta^*\rangle + \epsilon_{ji}$) is set to be a standard Gaussian random variable. We set the true parameter $\theta^*$ to be $s$-sparse where all the entries are zero except that the first $s$ entries are i.i.d random variables from a uniform distribution in [0,1]. Under both two models, we set the hard thresholding parameter $k$ greater than s but less than $3 s$. 
\vspace{-0.02in}

Here we compare the algorithms in different settings of $(n,d,m,s)$ and plot the estimation error $\|\theta^h - \theta^*\|_2$ over rounds of communications. The results of sparse linear regression and sparse logistic regression are showed in Figure~\ref{figure1} and Figure~\ref{figure2}. We can observe from these plots that: 

\begin{itemize}
	\item First, there is indeed a large gap between the local estimation error and the centralized estimation error. The estimation errors of EDSL and the Two-way Truncation decrease to the centralized one in the first several rounds of communications.

	\item Second, the Two-way Truncation algorithm is competitive with EDSL in both statistical accuracy and convergence rate as the theory indicated. Since it can converge in at least the same speed as EDSL's and requires less communication and computation cost in each iteration, overall it's more communicationally and computationally efficient.   
	%	\vspace{1in}
\end{itemize}
The above results support the theory that the Two-way Truncation approach is indeed more efficient and competitive to the centralized approach and EDSL.
\vspace{-0.05in}
%$\cO(\sqrt{\frac{s}{n}})$   
%\vspace{0.2 in}

\vspace{-0.05in}

\subsection{Real data}
\begin{figure}[!htb]
	\vspace{-0.1in}
	\begin{center}
		\hspace{-0.08in}\subfloat[dna (linear regression)]{
			\includegraphics[width=0.46\linewidth]{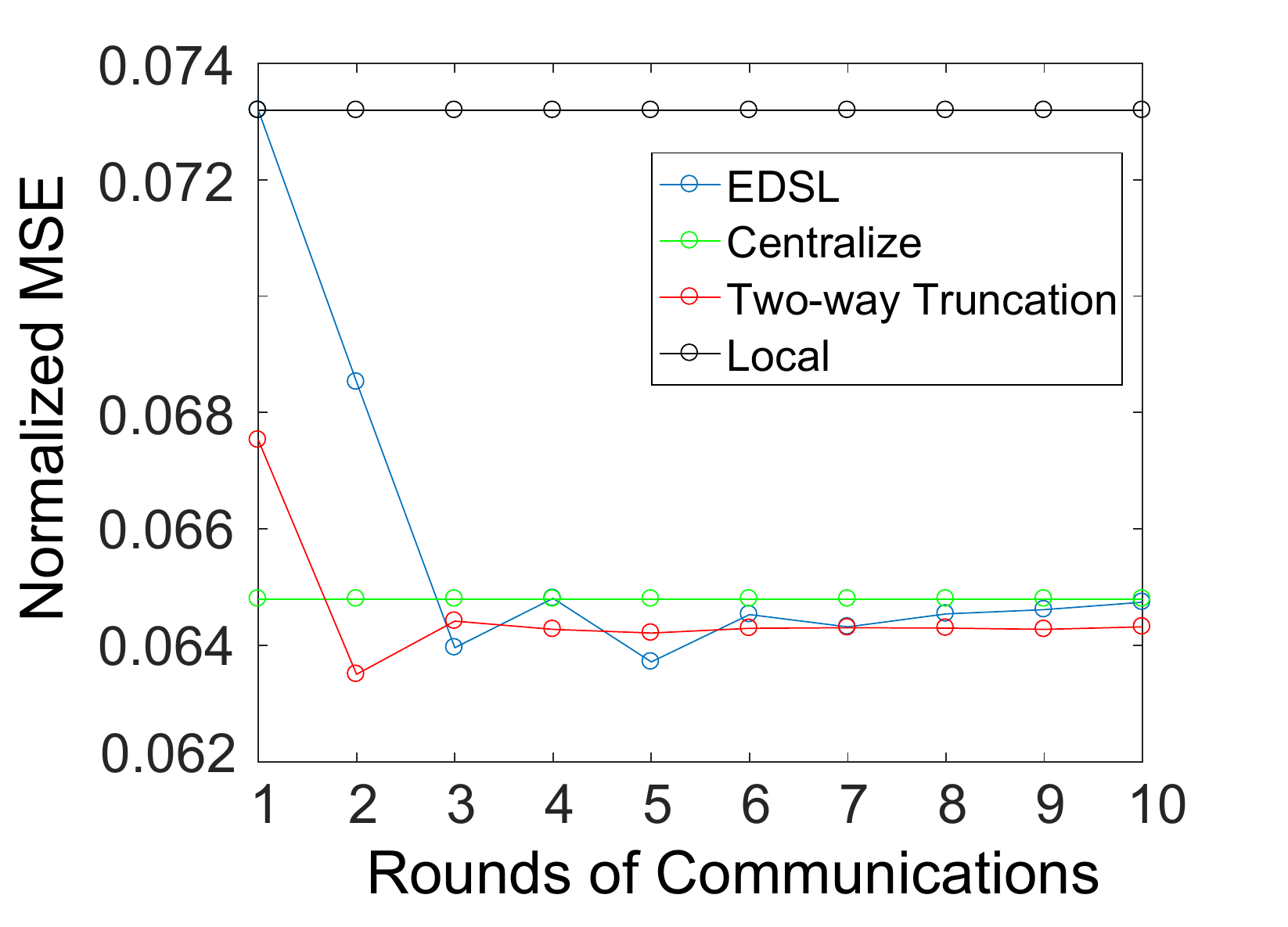}
		}
		\hspace{-0.08in}\subfloat[a9a (classification)]{
			\includegraphics[width=0.46\linewidth]{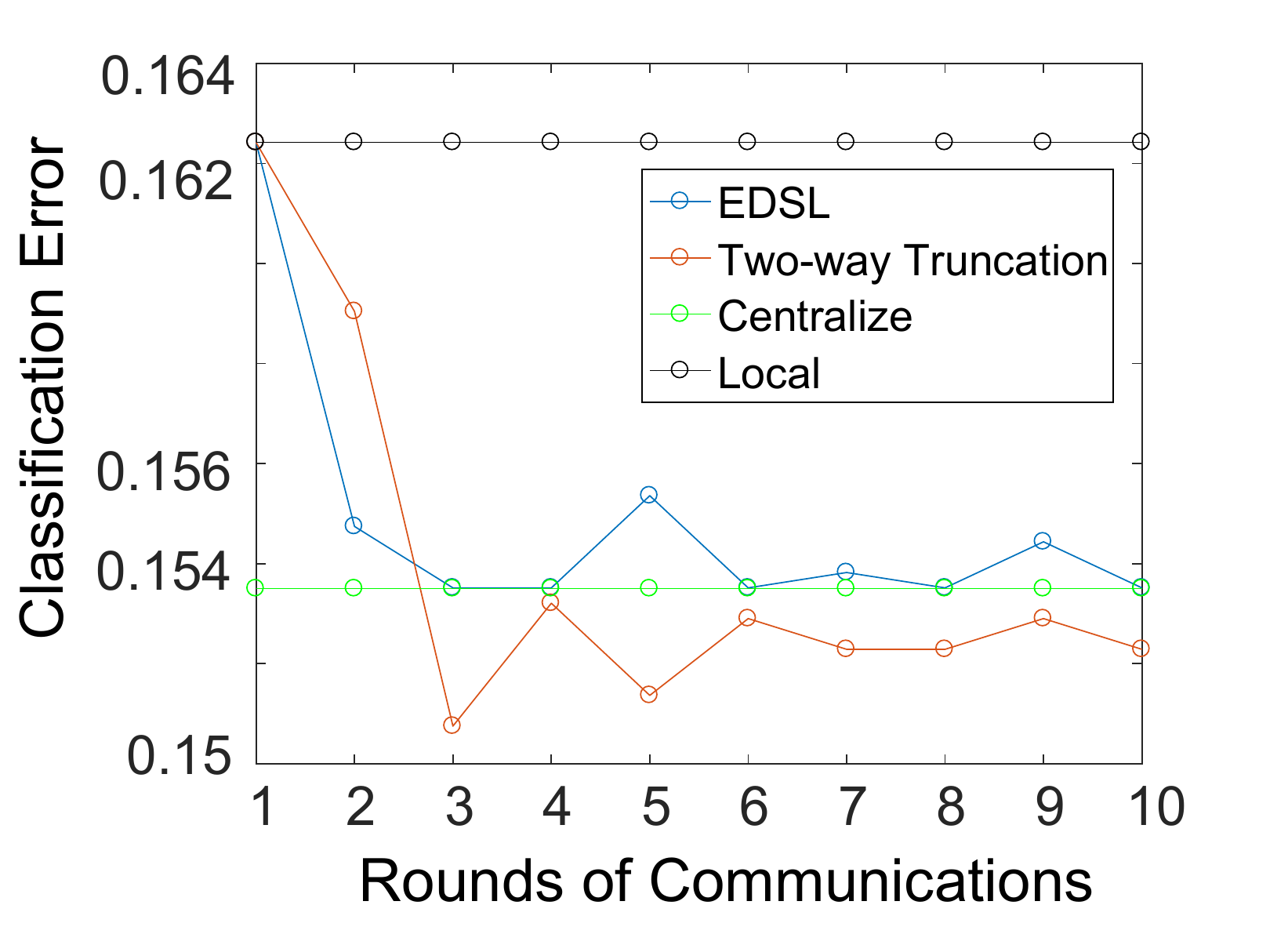}
		}
		%		\hspace{-0.096in}\subfloat[OP]{
		%			\includegraphics[width=0.45\linewidth]{{plot/OP_outlier_PaviaU}.pdf}
		%		}\hspace{-0.11in}
	\end{center}
	\vspace{-0.1in}
	\caption{Comparison among four algorithms on real datasets}\label{fig:real_data1}   
	\vspace{-0.5in}
\end{figure}
%\vspace{-0.5 in}
\vspace{0.4in}
In this section, we examine the above sparse learning algorithms on real-world datasets. The data comes from UCI Machine Learning Repository \footnote{http://archive.ics.uci.edu/ml/} and the LIBSVM website \footnote{https://www.csie.ntu.edu.tw/$\sim$cjlin/libsvmtools/datasets/}. The high-dimensional data 'dna' and 'a9a' are used in the regression model and classification model respectively. We randomly partition the data in $[60\%, 20\%,20\%]$ for training, validation and testing respectively. Here the data is divided randomly on $m = 10$ machines and processed by algorithms mentioned above. The results are summarized in Figure \ref{fig:real_data1}. These results in real-world data experiments again validate the theoretical analysis that the proposed Two-way Truncation approach is a quite effective sparse learning method with very small communication and computation costs. 

\section{Conclusions}

In this paper we propose a novel distributed sparse learning algorithm with Two-way Truncation.  Theoretically, we prove that the algorithm gives an estimation that converges to the minimizer of the expected loss exponentially and attain nearly the same statistical accuracy as EDSL and the centralized method. Due to the truncation procedure, this algorithm is more efficient in both communication and computation. Extensive experiments on both simulated data and real data verify this statement.

% use section* for acknowledgement
\section*{Acknowledgment}

The authors graciously acknowledge support from NSF Award CCF-1217751 and DARPA Young Faculty Award N66001-14-1-4047 and thank Jialei Wang for very useful suggestion. 

\bibliographystyle{CADSLTT}
\bibliography{CADSLTT}

% trigger a \newpage just before the given reference
% number - used to balance the columns on the last page
% adjust value as needed - may need to be readjusted if
% the document is modified later
%\IEEEtriggeratref{8}
% The "triggered" command can be changed if desired:
%\IEEEtriggercmd{\enlargethispage{-5in}}

% references section

% can use a bibliography generated by BibTeX as a .bbl file
% BibTeX documentation can be easily obtained at:
% http://www.ctan.org/tex-archive/biblio/bibtex/contrib/doc/
% The IEEEtran BibTeX style support page is at:
% http://www.michaelshell.org/tex/ieeetran/bibtex/
%\bibliographystyle{IEEEtran}
% argument is your BibTeX string definitions and bibliography database(s)
%\bibliography{IEEEabrv,../bib/paper}
%
% <OR> manually copy in the resultant .bbl file
% set second argument of \begin to the number of references
% (used to reserve space for the reference number labels box)
%\begin{thebibliography}{1}
%
%%\bibitem{IEEEhowto:kopka}
%%H.~Kopka and P.~W. Daly, \emph{A Guide to \LaTeX}, 3rd~ed.\hskip 1em plus
%%  0.5em minus 0.4em\relax Harlow, England: Addison-Wesley, 1999.
%\end{thebibliography}

% that's all folks
\end{document}